\documentclass[10pt,twocolumn,letterpaper]{article}

\usepackage{cvpr}              % To produce the CAMERA-READY version
\usepackage[dvipsnames]{xcolor}

\definecolor{cvprblue}{rgb}{0.21,0.49,0.74}
\usepackage[pagebackref,breaklinks,colorlinks,citecolor=cvprblue]{hyperref}

\def\paperID{6} % *** Enter the Paper ID here
\def\confName{CVPR}
\def\confYear{2024}

\title{LLMGeo: Benchmarking Large Language Models on \\ Image Geolocation In-the-wild}

\author{
Zhiqiang Wang\textsuperscript{1*}, Dejia Xu\textsuperscript{2*}, Rana Muhammad Shahroz, Khan\textsuperscript{3}, Yanbin Lin\textsuperscript{1}, Zhiwen Fan\textsuperscript{2} \and
Xingquan Zhu\textsuperscript{1+}\\
\\
\textsuperscript{1}{\small Florida Atlantic University,} \textsuperscript{2}{\small University of Texas at Austin,} \textsuperscript{3}{\small Vanderbilt University} \\
\textsuperscript{*}{\small Equal Contribution,} \textsuperscript{+}{\small Corresponding Author}\\
{\small https://github.com/yeyimilk/LLMGeo}
}

\begin{document}

\maketitle

\thispagestyle{plain} % Ensure page number on the first page

\begin{abstract}
% The ABSTRACT is to be in fully justified italicized text, at the top of the left-hand column, below the author and affiliation information.
% Use the word ``Abstract'' as the title, in 12-point Times, boldface type, centered relative to the column, initially capitalized.
% The abstract is to be in 10-point, single-spaced type.
% Leave two blank lines after the Abstract, then begin the main text.
% Look at previous \confName abstracts to get a feel for style and length. 

Image geolocation is a critical task in various image-understanding applications. However, existing methods often fail when analyzing challenging, in-the-wild images.
Inspired by the exceptional background knowledge of multimodal language models, we systematically evaluate their geolocation capabilities using a novel image dataset and a comprehensive evaluation framework. We first collect images from various countries via Google Street View. Then, we conduct training-free and training-based evaluations on closed-source and open-source multi-modal language models. we conduct both training-free and training-based evaluations on closed-source and open-source multimodal language models. Our findings indicate that closed-source models demonstrate superior geolocation abilities, while open-source models can achieve comparable performance through fine-tuning.

% The recent success of multimodal language models with 
% Code and dataset are available in GitHub: https://github.com/yeyimilk/LLMGeo.git
\end{abstract}    

\begin{figure*}[th]
  \centering
  \includegraphics[width=\textwidth]{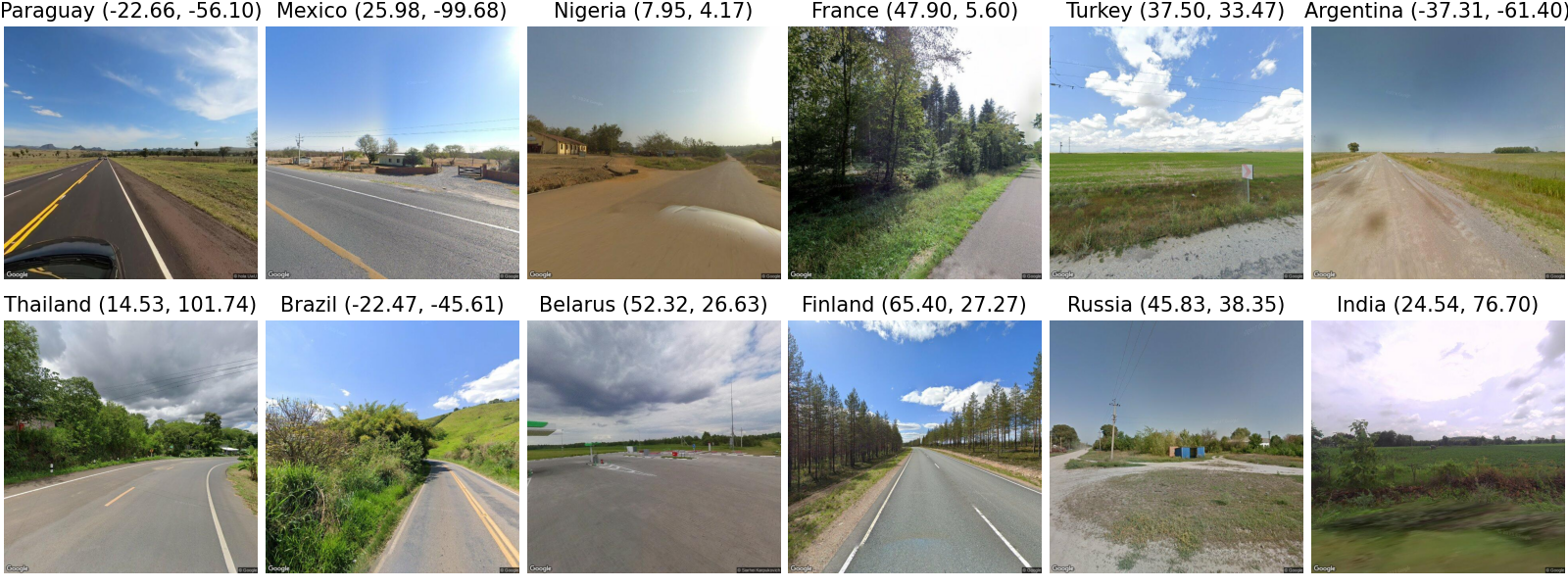}
  \caption{Image samples from the test set.}
    \label{fig:samples}
\end{figure*}

\begin{figure*}[th]
  \centering
  \includegraphics[width=\textwidth]{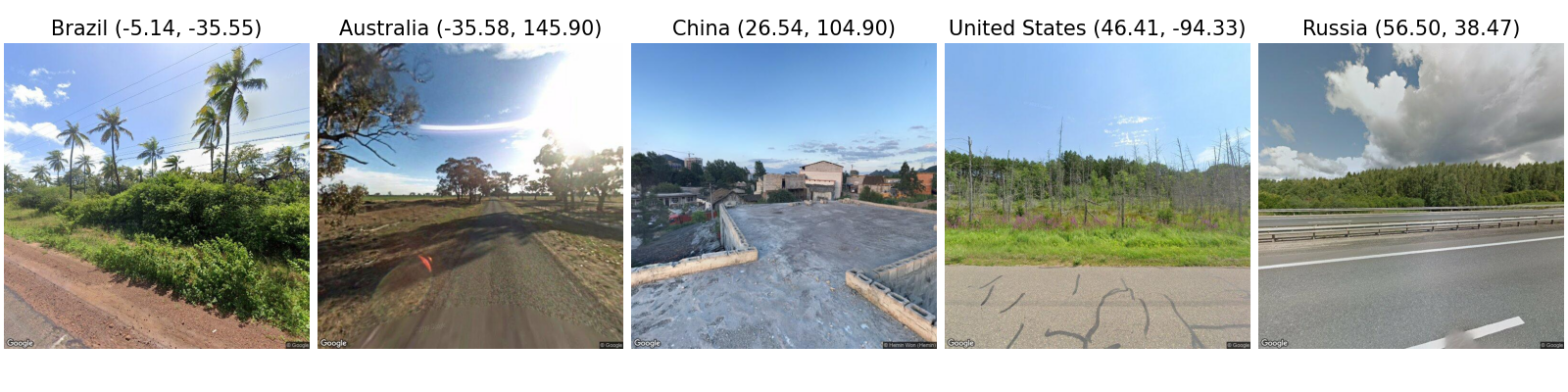}
  \caption{The five images are used as fixed input, including their order, for the static few shots strategy.}
    \label{fig:static_shots}
\end{figure*}

\begin{figure*}[h]
  \centering
  \includegraphics[width=\textwidth]{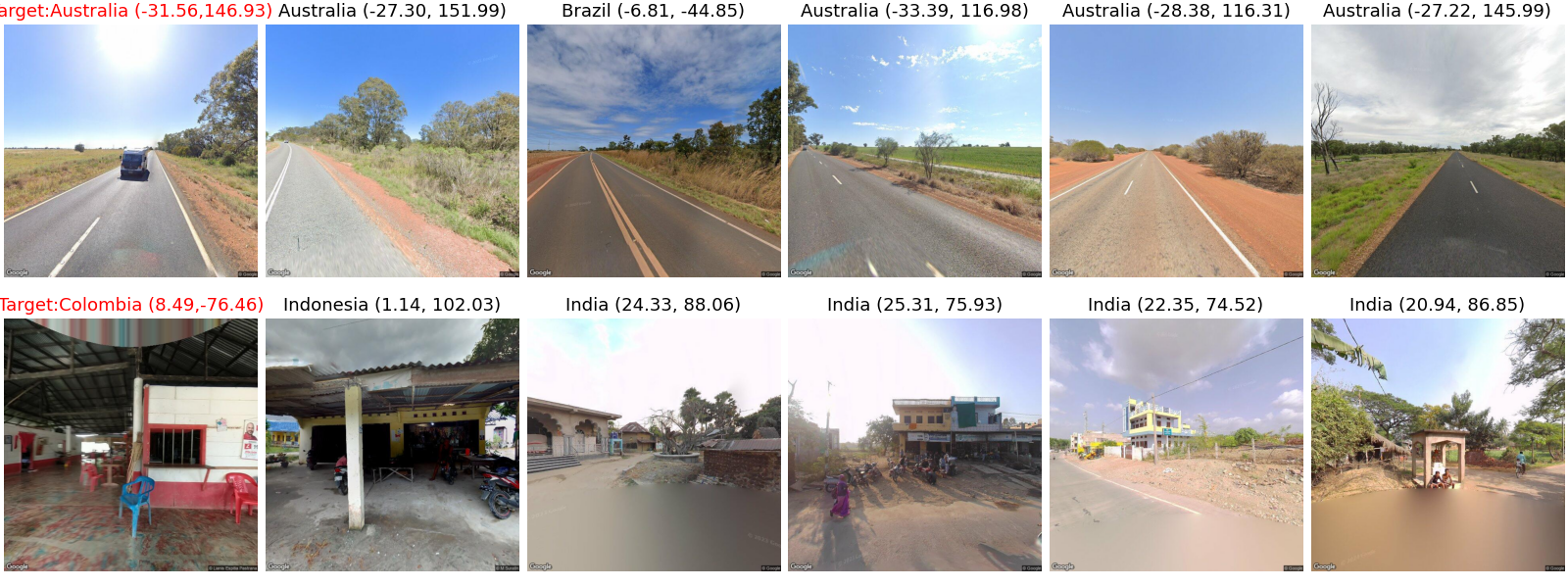}
  \caption{Images samples about dynamic few shots strategy. The first image is the target image, which is for LLMs to guess where it was taken, and the following images on the same row are their corresponding five most similar images based on CLIP embeddings ordered by Euclidean distance descending.}
    \label{fig:dynamic_shots}
\end{figure*}

\section{Introduction}
\label{sec:intro}

% Importance of geolocation

% Previous works

% AI safety.. privacy concerns

% LLM 

% Our benchmark

Image geolocation refers to the process of determining the specific geographic location from which a given image was taken. This geographic information is crucial across various domains, including urban planning, environmental monitoring, and social media analysis. The ability to automatically identify the location of images provides valuable insights and supports numerous applications, such as augmented reality, location-based services, and geotagging.

Despite its importance, image geolocation in the wild remains a challenging task, particularly when dealing with images sourced from diverse sources such as social media platforms and online repositories. While many previous works~\cite{haas2023pigeon,hays2008im2gps,wu2022im2city,suresh2018deepgeo,cepeda2023geoclip} have developed curated loss functions and model designs to tackle this challenge, they usually have compromised performance when evaluated on images in the wild. 
In contrast, recent advances in large multimodal models (LMMs) have demonstrated impressive capabilities in background knowledge across a broad range of tasks. These models are trained on large-scale datasets, exhibiting outstanding understanding~\cite{yue2023mmmu, wang2023large}, reasoning~\cite{lu2023mathvista}, and commonsense~\cite{zhang2022visual} abilities.

% Previous efforts in this domain have primarily focused on traditional computer vision techniques, often relying on handcrafted features and limited contextual information.

% Furthermore, large language models (LLMs) have opened up new avenues for advancing geolocation capabilities. These models, equipped with both textual and visual understanding capabilities, offer the potential to surpass the performance of traditional methods by leveraging vast amounts of multimodal data.

% Since the 
% There is growing concern
% Despite the exciting progress, 
While numerous benchmarks have been established to evaluate various image understanding abilities of multimodal language~\cite{bavishi2024fuyu,dong2024internlm,liu2023llava,liu2023improvedllava}, 
little attention has been paid to their geolocation capabilities. To address this research gap, we conduct the first systematic analysis of image geolocation abilities. First, we introduce a large-scale dataset of in-the-wild images sampled from diverse geolocations. Second, we comprehensively benchmark the capabilities of both open-source and closed-source multimodal language models through training-free and training-based evaluations.

% To this end, we conduct the first systematic analysis of image geolocation ability to fill the 
% research gap. Firstly, we present a large-scale dataset of in-the-wild images sampled from diverse geolocations. Secondly, we comprehensively benchmark the abilities of existing open-source and closed-source multi-modal language models through training-free and training-based evaluations. 
% Our findings reveal the ability of 
% construct a comprehensive 
% In this study, we conduct the first systematic analysis of image geolocation abili
% of these multimodal language models.
% though geolocation 
% which is a key building block of 
% However, integrating LLMs with vision capabilities into geolocation tasks presents unique challenges and opportunities. Concerns about AI safety and privacy have emerged, underscoring the need for robust and ethically sound approaches to geolocation. Additionally, evaluating LLMs' performance in geolocation tasks requires comprehensive benchmarking frameworks that account for various model architectures, datasets, and evaluation metrics.

% \textit{Contributions:}
Our contributions can be summarized as follows,
\begin{itemize}
    \item \textbf{Introduction of a novel image dataset:} We present a new dataset, exclusively sourced from Google Street View, designed to challenge Large Multimodal Models (LMMs) through real-world, in-the-wild random images. This dataset is intended to serve as a robust benchmark for assessing these models' ability to identify image locations accurately.
    \item \textbf{Comprehensive evaluation framework:} We evaluate a diverse set of LMMs, including state-of-the-art closed-source models like GPT-4V and Google Gemini, and promising open-source models such as BLIP~\cite{li2023blip}, Fuyu~\cite{bavishi2024fuyu}, InternLM-VL~\cite{dong2024internlm}, and LLaVA~\cite{liu2023llava,liu2023improvedllava}. Our evaluations, both training-free and training-based, thoroughly assess these models' geolocation accuracies at the country level and their adaptability to challenging in-the-wild image data.

\end{itemize}

% Overall, "LLMGeo" not only benchmarks the current landscape of LLM capabilities in geolocation but also sets the stage for future advancements in the field by highlighting areas for improvement and proposing a path forward for responsible and effective use of AI technologies.

\section{Dataset}
\label{sec:dataset}

\begin{table*}[h]
\centering
\caption{\small{Image distribution and orientations from the proposed dataset.}}
\begin{tabular}{lcccccc}
\toprule
& H-0 & H-90 & H-180 & H-270 & Total & \# of Countries \\
\midrule
Test                      & 250  & 250  & 250  & 250  & 1000 & 82 \\
Train                     & 618  & 600  & 600  & 600  & 2418 & 92 \\
Comprehensive Train       & 1609 & 1600 & 1599 & 1598 & 6408 & 115 \\
\bottomrule
\multicolumn{4}{l}{\footnotesize H: the compass heading of the camera.} \\

\end{tabular}
\label{tab:img_distribution}
\end{table*}

\begin{table*}[h]
\centering
\caption{\small{Distance pairs(km)}}
\renewcommand{\arraystretch}{1.3}
\begin{tabular}{lcccccc}
\toprule
&  $d < 10$ & $10 \leqslant d < 100$ & $100 \leqslant d < 500$ & $500 \leqslant d < 1000$ & $ 1000 \leqslant d$ \\
\midrule
Test                      & 0.00012\% & 0.07\%  & 1.13\%  & 2.42\% & 96.38\%  \\
Train                     & 0.0008\%  & 0.06\%  & 1.11\%  & 2.41\% & 96.41\%  \\
Comprehensive Train       & 0.0062\%  & 0.008\% & 1.19\%  & 2.64\% & 96.09\% \\
\bottomrule

\end{tabular}
\label{tab:distance_distribution}
\end{table*}

Our dataset's images are directly from Google Street View, while specific parameters were set in the API request to mimic common human sight. The up or down angle of the camera relative to the Street View vehicle is set to 0 degrees to maintain a natural, level perspective. The horizontal field of view is fixed at 90 degrees, mirroring the horizontal scope typical of human vision. To capture diverse viewpoints, the camera's compass heading is adjusted to four fixed orientations: 0 (North), 90 (East), 180 (South), and 270 (West) degrees. All the images have the same size of 512x512. 

Figure \ref{fig:samples} displays random sample images from our test set, which predominantly consists of natural landscapes and rural scenes with features such as water bodies, trees, and agricultural fields. These images do not include prominent urban infrastructure or significant man-made constructs, thereby increasing the complexity and intrigue of identifying each image's geographical origin.

Table \ref{tab:img_distribution} provides a statistical overview of the dataset, detailing the distribution of these varied perspectives. It shows a methodical approach to capturing diverse orientations and geographic locations within the proposed dataset. The table divides the dataset into three subsets: Test, Train, and Comprehensive Train, each detailed with the count of images across four compass headings and the total number of images alongside the number of represented countries. From the perspective of camera headings, the dataset maintains a remarkable balance across all subsets, with each heading represented almost equally. The test set, train set, and comprehensive train set have 1000, 2418, and 6408 images, with each heading having exactly or around one-quarter of the total images. When only considering the countries that appear in the test set, there are 2388 and 6011 images in the train and comprehensive train set, respectively.

\subsection{Dataset Distance Pairs Analysis}

Larger countries are more likely to have more images, so a strategy was taken to give bigger countries more chances during the random image pick-up process. This results in a significant imbalance in the representation of countries within the dataset. For instance, a country might have as few as one image in the dataset. In contrast, another could have as many as 121, 293, and 661 images in the test, training, and comprehensive training sets. To mitigate the potential impact of this geographic imbalance on model training and evaluation, we implemented a policy ensuring that each country represented in the Test set is also represented in the Train set with at least two images and in the Comprehensive Train set with at least four images. 

Table \ref{tab:distance_distribution} shows the geographical diversity within our dataset. We analyzed the physical distances between image pairs based on their geolocations, categorizing them into intervals ranging from less than 10 kilometers(km) to over 1000 km. The result reveals a strategic emphasis on maximizing geographical variance, with most image pairs—more than 96\% across Test, Train, and Comprehensive Train sets—showing separations of over 1000 kilometers. The distribution significantly reduces the probability of selecting visually similar images from proximal locations, ensuring the dataset spans a broad spectrum of environmental and urban landscapes. 
\section{Experiment Settings}
\label{sec:experiment}

\begin{table*}[th]
\centering
\caption{\small{Training-free evaluation results in different scenarios.}}
\begin{tabular}{lccccccc}
\toprule
& Basic & Must & Tips & S-5-shot  & D-5-shot & S-5-shot-Rd & D-5-shot-Rd \\
\midrule
GeoCLIP                     & 0.258 & - & - & - & - & - & -   \\
\hline
GPT-4V                  & 0.102 & 0.513 & 0.422 & - & - & - & -  \\
Gemini                      & \textbf{0.666} &\textbf{ 0.660} & \textbf{0.670 }& \textbf{0.741}  & \textbf{0.736} & \textbf{0.737} & \textbf{0.746}   \\
\hline
BLIP-2-2.7B                 & 0.290 & 0.305 & 0.002 & - & - & - & -     \\
BLIP-2-T5-XL                & 0.257 & 0.365 & 0.361 & - & - & - & -    \\
Fuyu-8B                     & 0.014 & 0.016 & 0.008 & - & - & - & -   \\
ILM-VL-7B                   & 0.182 & 0.301 & 0.327 & 0.000 & 0.016 & 0.024 & 0.015    \\
LLaVA1.5-7B                 & 0.189 & 0.204 & 0.120  & 0.027 & 0.317 & 0.031 & 0.321  \\
LLaVA1.5-13B                & 0.165 & 0.185 & 0.049 & 0.032 & 0.310 & 0.035 & 0.312 \\
\bottomrule
\end{tabular}
\label{tab:accuracy}
\end{table*}

% \subsection{Evaluated Models}

%  Two close source LLMs, GPT-4V and Google Gemini, and four open source LMMs, BLIP-2~\cite{li2023blip}, Fuyu~\cite{bavishi2024fuyu}, InternLM-VL(IML-VL)~\cite{dong2024internlm}, LLaVA1.5~\cite{liu2023improvedllava} are employed for a thorough benchmark. A state-of-the-art image-to-GPS specialized model, GeoGLIP~\cite{cepeda2023geoclip}, is also included as a reference. The k-nearest neighbors (k-NN) algorithm was used for the dynamic few-shots strategy to get the five most similar images as examples in the input from the train set. More details can be found in Section \ref{x_suppl_models} and \ref{x_suppl_d_few_shots_s}.

\subsection{Experiment Models} \label{x_suppl_models}

In this section, we elaborate on the models that we evaluate.

\begin{itemize}
    \item \textbf{GeoCLIP\cite{cepeda2023geoclip}:} A groundbreaking approach inspired by CLIP for Image-to-GPS retrieval, designed to enhance the alignment between images and their corresponding GPS coordinates.
    
    \item \textbf{ChatGPT-4V\cite{OpenAIGPT4V2023}:} An extension of the ChatGPT model with integrated visual processing capabilities, enabling it to understand and generate content based on text and images.
    
    \item \textbf{Gemini\cite{team2023gemini}: } Gemini introduces a versatile multimodal model family excelling in understanding across images, audio, video, and text, with its vision capabilities setting new benchmarks in image-related tasks and multimodal reasoning.
    
    \item \textbf{Blip-2\cite{li2023blip}: } BLIP-2 introduces a cost-effective vision-language pre-training approach that leverages existing pre-trained models with a Querying Transformer, achieving state-of-the-art results in vision-language tasks with significantly fewer trainable parameters.
    
    \item \textbf{Fuyu\cite{bavishi2024fuyu}:} Fuyu stands out with its simple yet versatile architecture, excelling in digital agent tasks and offering rapid, high-resolution image processing capabilities.
    
    \item \textbf{InternLM-XComposer2 (ILM-VL)\cite{dong2024internlm}: } It innovates in vision-language interaction with a Partial LoRA technique, excelling in creating and understanding complex text-image content, setting new benchmarks in multimodal performance.
    
    \item \textbf{LlaVA\cite{liu2024llavanext}:} LLaVA 1.5 sets a new standard in large multimodal models with a highly efficient vision-language connector, achieving unprecedented performance on 11 benchmarks using minimal data and training resources.
    
\end{itemize}

\subsection{Dynamic few-shots strategy} \label{x_suppl_d_few_shots_s}

For the dynamic few-shots strategy, derived from Retrieval-Augmented Generation(RAG) techniques\cite{lewis2020retrieval}, DINOv2\cite{oquab2023dinov2} and CLIP\cite{radford2021learning} were employed to generate embedding features from the train and test set. After that, for each image in the test set, the kNN algorithm was used to find similar images from the train set. Table \ref{tab:knn_embeddings} shows that CLIP outperforms DINOv2 in the top 1 and top 5 evaluation levels, achieving an accuracy of 0.312 and 0.586, respectively. When LLMs were evaluated with the dynamic few-shots strategy, for each image to be guessed, the top 5 images were determined by the kNN algorithm through embeddings generated by CLIP as it has better results than DINOv2.

\begin{table}[h]
\centering
\caption{\small{kNN results for test set within train set with embedding feature vectors}}
\renewcommand{\arraystretch}{1.3}
\begin{tabular}{lcc}
\toprule
& Top 1 & Top 5 \\
\midrule
DINOv2                   & 0.281 & 0.539  \\
CLIP                     & 0.313 & 0.586 \\
\bottomrule
\end{tabular}
\label{tab:knn_embeddings}
\end{table}

\subsection{Prompt Strategies}

In this section, we elaborate on the prompting strategies used for evaluations.

\begin{itemize}
    \item \textbf{Basic:} The model is shown an image and prompted to guess the country where the image was captured, relying solely on visual cues present. With this strategy, LLMs prefer responding to the "unknown" when the image is not easily identified.
    \item \textbf{Must:} To address cases where limited information may prevent answering a country, we employ imperative prompts to compel the model to make a country guess for each image.
    \item \textbf{Tips: } We offer general guidelines to the model, suggesting it consider factors like sun position, license plates, and other identifiable features within the image to infer the geographic location without directly providing this specific information. These uniform guidelines apply to all models across every evaluation round.
    \item \textbf{S-5-shot: } The model is given five additional images, each tagged with their respective countries, as references before it predicts the country of a new image. These reference images remain consistent across all models and evaluation rounds. An example is shown in Figure \ref{fig:static_shots}.
    \item \textbf{D-5-shot: } Similar to the S-5-shot method, but the five reference images are specifically chosen based on their proximity to the target image, utilizing the k-Nearest Neighbors (kNN) algorithm from the training set based on their embeddings generated by CLIP\cite{radford2021learning}, and ranked by their closeness. Figure \ref{fig:dynamic_shots} shows two sets of example input images for this strategy.
    \item \textbf{S-5-shot-Rd:} Adapting the S-5-shot method, the order of the five reference images is randomized, challenging the model to identify relevant patterns without depending on the sequence.
    \item \textbf{D-5-shot-Rd:} Following the D-5-shot strategy, this method randomizes the order of the selected images, disregarding their proximity, to evaluate the model's ability to utilize non-sequential cues for geographic deduction.
\end{itemize}

% Note that not all models support multiple text input

\begin{table*}[th]
\centering
\caption{\small{Training-based evaluation results.}}
\begin{tabular}{@{\hspace{5mm}}l@{\hspace{4mm}}|@{\hspace{8mm}}c@{\hspace{10mm}}c@{\hspace{10mm}}c@{\hspace{5mm}}}
    \toprule
    & Basic ($\uparrow$) & Must ($\uparrow $) & Tips ($\uparrow $)\\
    \hline
    ILM-VL-7B(T)     & 0.413 (\textcolor{red}{+0.231}) & 0.436 (\textcolor{red}{+0.135}) & 0.449 (\textcolor{red}{+0.122})  \\
    ILM-VL-7B(CT)    & 0.441 (\textcolor{red}{+0.259}) & 0.443 (\textcolor{red}{+0.142}) & 0.439 (\textcolor{red}{+0.112})  \\
    LLaVA-7B (T)     & 0.562 (\textcolor{red}{+0.373}) & \textbf{0.561 (\textcolor{red}{+0.357})} & 0.547 (\textcolor{red}{+0.427})      \\
    LLaVA-7B (CT)    & 0.557 (\textcolor{red}{+0.368}) & 0.560 (\textcolor{red}{+0.356}) & \textbf{0.548 (\textcolor{red}{+0.428})} \\
    LLaVA-13B (T)    & \textbf{0.567 (\textcolor{red}{+0.402})} & 0.391 (\textcolor{red}{+0.206}) & 0.342 (\textcolor{red}{+0.293}) \\
    LLaVA-13B (CT)   & 0.562 (\textcolor{red}{+0.397}) & 0.385 (\textcolor{red}{+0.200}) & 0.329 (\textcolor{red}{+0.280}) \\
    \bottomrule
    \multicolumn{4}{l}{\footnotesize T: finetune with train set; CF: finetune with comprehensive train set} \\
\end{tabular}
\label{tab:finetune}
\end{table*}

\subsection{Training-free Evaluation}

Table \ref{tab:accuracy} shows the training-free evaluation results with different prompts input except for GeoGLIP, as it only takes the image as input, and its output is geolocation.

From Table \ref{tab:accuracy}, we can see that Gemini performs better than other models in all strategies. Gemini achieves similar accuracy, nearly $0.67$, for the Basic, Must, and Tips strategies. It also outperforms comparable models using the few shots strategies with an accuracy of up to $0.746$. We did not test few-shot scenarios for the ChatGPT-4V model, while the current BLIP and Fuyu do not support using multiple images as input.

In terms of open-source models, BLIP-2-2.7B has the highest accuracy for the Basic prompt, and BLIP-2-T5-XL achieves best for the Must and Tips prompt cases, with an accuracy of 0.365 and 0.361, respectively. The accuracy of the Tips case for model BLIP-2-2.7B drops to 0.002 because the model is very sensitive to the text input and unable to handle the context if it is relatively long.

The ILM-VL model achieves good performance in the single image input cases and for the few-shot cases; while the ILM-VL model can take a few images as input, its ability to deal with multiple images in question-and-answer tasks almost drops to zero. 

The few-shot strategies show their effectiveness for Gemini, while the static, dynamic, and random strategies do not significantly affect Gemini. As for LLaVA, taking 5 closest images with their country names as part of the prompt for the guessed image can significantly improve the accuracy by more than 50\% compared to the highest accuracy for only text input as prompt. Taking the same 5 images with their country names for every round of Q\&A tasks does hurt the performance. This can be attributed to the hyperparameter of temperature being set to 0. In this case, as the image to be guessed is only a small portion of the input, the output may be preferred to stick to similar outputs inherited from the inputs.  Finally, the 4 outcomes of the few-shot strategies also demonstrate that the input order of the input images only shows a minor impact on the accuracy.

\subsection{Training-based Evaluation}

Table \ref{tab:finetune} illustrates the efficacy of our dataset in enhancing the accuracy of LLMs for determining the location of images. The results indicate a significant improvement when models are fine-tuned with either the train set or a comprehensive train set, employing Basic, Must, and Tips strategies. The enhancement in accuracy, observed after fine-tuning models with our dataset, can be substantial—more than double in some cases. 

LLaVA-13B(T) has the highest accuracy, 0.567, along with the strategy of the Basic strategy. However, the outstanding performance is not significant as LLaVA-7B achieved an accuracy of around 55\% across three strategies and 2 train sets. It outperforms the close source model ChatGPT-4V in those three cases. ILM-VL also shows better results to above 40\% after fine-tuning, which surpasses all the open source models before fine-tuning.

One noticeable thing is that fine-tuning an LLM with more images along with an answer only does not guarantee better performance in this geolocation guessing task. It can be observed that there are 6 of 9 cases in the model where fine-tuning with the train set shows higher performance than fine-tuning with the comprehensive train set.

% \subsection{Training-free Evaluation}

% Table \ref{tab:accuracy} presents the accuracy of various LLMs in identifying the country where an image was taken based on different prompt strategies before any fine-tuning. Gemini emerges as the most accurate model across all tested strategies. Regarding open-source models, the accuracy can be as high as 0.327 by InternLM-VL and as low as 0.008 by Fuyu-8B. The few-shot prompting strategies show an extremely negative impact on those open-source models, suggesting that these models are not well-designed for reasoning on inputs containing multiple images. 
% More details and analysis can be found in Supplementary (Section~\ref{x_suppl_training_free_eva}).

% \subsection{Training-based Evaluation}

% Table \ref{tab:finetune} shows significant improvements in accuracy for identifying the country from images across basic, must, and tips scenarios after finetuning. The LLaVA models, particularly the 7B variants, exhibit the most substantial gains, with accuracy increases exceeding 0.35 in multiple scenarios, demonstrating the effectiveness of the finetuning process. 
% Note that after fine-tuning, the open-source models achieve on-par or superior performance compared to GPT-4V, while there is still a gap when compared with Gemini's abilities.
% % , these findings suggest that 
% More details can be found in Supplementary(Section \ref{x_suppl_training_based_eva}).
\section{Discussion}
\label{sec:discussion}

In this work, we conduct the first systematic study in image geolocation abilities of multimodal language models. We first introduce a novel dataset comprised of images sampled from Google Street View API. The dataset is diverse, encompassing varied perspectives and landscapes from multiple countries, which allows for comprehensive benchmarking of multimodal language mnodels' geolocation abilities. We employed multiple training-free evaluation strategies from simple prompts, chain-of-thought, and few shot prompting.  We further fine-tuned two open-source models using our collected dataset, which significantly enhanced the accuracy of these models in predicting the geographic origin of the images at the country level.

While our findings contribute valuable insights into the capabilities of LLMs in image-based geolocation tasks, several limitations are notable. Firstly, our evaluations were confined to country-level geolocation without extending it to more granular levels, such as state and city identifications. Additionally, the majority of our dataset images are natural landscapes and rural scenes, which may not adequately represent the complexity and diversity of urban environments.

In future research, we aim to test geolocation accuracy at more granular levels or even provide a precise latitude and longitude coordinate. This expansion will allow us to understand better LLMs' capabilities and limitations in more densely populated and geographically complex environments. Furthermore, to address the current dataset’s emphasis on natural and rural landscapes, we plan to enrich it with a broader array of images, including urban settings with diverse architectural styles and infrastructural elements. This enhancement will provide a more robust LLM testbed and potentially improve the models' usefulness in practical, real-world applications where urban geolocation is critical.

% \section{Discussion}
% Summary 

% Limitations and Future work

% \input{sec/2_formatting}
% \input{sec/3_finalcopy}
{
    \small
    \bibliographystyle{ieeenat_fullname}
    \bibliography{main}

\begin{thebibliography}{20}
\providecommand{\natexlab}[1]{#1}
\providecommand{\url}[1]{\texttt{#1}}
\expandafter\ifx\csname urlstyle\endcsname\relax
  \providecommand{\doi}[1]{doi: #1}\else
  \providecommand{\doi}{doi: \begingroup \urlstyle{rm}\Url}\fi

\bibitem[Bavishi et~al.(2024)Bavishi, Elsen, Hawthorne, Nye, Odena, Somani, and Ta{\c{s}}{\i}rlar]{bavishi2024fuyu}
Rohan Bavishi, Erich Elsen, Curtis Hawthorne, Maxwell Nye, Augustus Odena, Arushi Somani, and Sa{\u{g}}nak Ta{\c{s}}{\i}rlar.
\newblock Fuyu-8b: A multimodal architecture for ai agents, 2024.

\bibitem[Cepeda et~al.(2023)Cepeda, Nayak, and Shah]{cepeda2023geoclip}
Vicente~Vivanco Cepeda, Gaurav~Kumar Nayak, and Mubarak Shah.
\newblock Geoclip: Clip-inspired alignment between locations and images for effective worldwide geo-localization.
\newblock \emph{arXiv preprint arXiv:2309.16020}, 2023.

\bibitem[Dong et~al.(2024)Dong, Zhang, Zang, Cao, Wang, Ouyang, Wei, Zhang, Duan, Cao, et~al.]{dong2024internlm}
Xiaoyi Dong, Pan Zhang, Yuhang Zang, Yuhang Cao, Bin Wang, Linke Ouyang, Xilin Wei, Songyang Zhang, Haodong Duan, Maosong Cao, et~al.
\newblock Internlm-xcomposer2: Mastering free-form text-image composition and comprehension in vision-language large model.
\newblock \emph{arXiv preprint arXiv:2401.16420}, 2024.

\bibitem[Haas et~al.(2023)Haas, Alberti, and Skreta]{haas2023pigeon}
Lukas Haas, Silas Alberti, and Michal Skreta.
\newblock Pigeon: Predicting image geolocations.
\newblock \emph{arXiv preprint arXiv:2307.05845}, 2023.

\bibitem[Hays and Efros(2008)]{hays2008im2gps}
James Hays and Alexei~A Efros.
\newblock Im2gps: estimating geographic information from a single image.
\newblock In \emph{2008 ieee conference on computer vision and pattern recognition}, pages 1--8. IEEE, 2008.

\bibitem[Lewis et~al.(2020)Lewis, Perez, Piktus, Petroni, Karpukhin, Goyal, K{\"u}ttler, Lewis, Yih, Rockt{\"a}schel, et~al.]{lewis2020retrieval}
Patrick Lewis, Ethan Perez, Aleksandra Piktus, Fabio Petroni, Vladimir Karpukhin, Naman Goyal, Heinrich K{\"u}ttler, Mike Lewis, Wen-tau Yih, Tim Rockt{\"a}schel, et~al.
\newblock Retrieval-augmented generation for knowledge-intensive nlp tasks.
\newblock \emph{Advances in Neural Information Processing Systems}, 33:\penalty0 9459--9474, 2020.

\bibitem[Li et~al.(2023)Li, Li, Savarese, and Hoi]{li2023blip}
Junnan Li, Dongxu Li, Silvio Savarese, and Steven Hoi.
\newblock Blip-2: Bootstrapping language-image pre-training with frozen image encoders and large language models.
\newblock In \emph{International conference on machine learning}, pages 19730--19742. PMLR, 2023.

\bibitem[Liu et~al.(2023{\natexlab{a}})Liu, Li, Li, and Lee]{liu2023improvedllava}
Haotian Liu, Chunyuan Li, Yuheng Li, and Yong~Jae Lee.
\newblock Improved baselines with visual instruction tuning, 2023{\natexlab{a}}.

\bibitem[Liu et~al.(2023{\natexlab{b}})Liu, Li, Wu, and Lee]{liu2023llava}
Haotian Liu, Chunyuan Li, Qingyang Wu, and Yong~Jae Lee.
\newblock Visual instruction tuning, 2023{\natexlab{b}}.

\bibitem[Liu et~al.(2024)Liu, Li, Li, Li, Zhang, Shen, and Lee]{liu2024llavanext}
Haotian Liu, Chunyuan Li, Yuheng Li, Bo Li, Yuanhan Zhang, Sheng Shen, and Yong~Jae Lee.
\newblock Llava-next: Improved reasoning, ocr, and world knowledge, 2024.

\bibitem[Lu et~al.(2023)Lu, Bansal, Xia, Liu, Li, Hajishirzi, Cheng, Chang, Galley, and Gao]{lu2023mathvista}
Pan Lu, Hritik Bansal, Tony Xia, Jiacheng Liu, Chunyuan Li, Hannaneh Hajishirzi, Hao Cheng, Kai-Wei Chang, Michel Galley, and Jianfeng Gao.
\newblock Mathvista: Evaluating mathematical reasoning of foundation models in visual contexts.
\newblock \emph{arXiv preprint arXiv:2310.02255}, 2023.

\bibitem[{OpenAI}(2023)]{OpenAIGPT4V2023}
{OpenAI}.
\newblock {GPT-4V(ision) System Card}, 2023.
\newblock Accessed: 2024-03-31.

\bibitem[Oquab et~al.(2023)Oquab, Darcet, Moutakanni, Vo, Szafraniec, Khalidov, Fernandez, Haziza, Massa, El-Nouby, et~al.]{oquab2023dinov2}
Maxime Oquab, Timoth{\'e}e Darcet, Th{\'e}o Moutakanni, Huy Vo, Marc Szafraniec, Vasil Khalidov, Pierre Fernandez, Daniel Haziza, Francisco Massa, Alaaeldin El-Nouby, et~al.
\newblock Dinov2: Learning robust visual features without supervision.
\newblock \emph{arXiv preprint arXiv:2304.07193}, 2023.

\bibitem[Radford et~al.(2021)Radford, Kim, Hallacy, Ramesh, Goh, Agarwal, Sastry, Askell, Mishkin, Clark, et~al.]{radford2021learning}
Alec Radford, Jong~Wook Kim, Chris Hallacy, Aditya Ramesh, Gabriel Goh, Sandhini Agarwal, Girish Sastry, Amanda Askell, Pamela Mishkin, Jack Clark, et~al.
\newblock Learning transferable visual models from natural language supervision.
\newblock In \emph{International conference on machine learning}, pages 8748--8763. PMLR, 2021.

\bibitem[Suresh et~al.(2018)Suresh, Chodosh, and Abello]{suresh2018deepgeo}
Sudharshan Suresh, Nathaniel Chodosh, and Montiel Abello.
\newblock Deepgeo: Photo localization with deep neural network.
\newblock \emph{arXiv preprint arXiv:1810.03077}, 2018.

\bibitem[Team et~al.(2023)Team, Anil, Borgeaud, Wu, Alayrac, Yu, Soricut, Schalkwyk, Dai, Hauth, et~al.]{team2023gemini}
Gemini Team, Rohan Anil, Sebastian Borgeaud, Yonghui Wu, Jean-Baptiste Alayrac, Jiahui Yu, Radu Soricut, Johan Schalkwyk, Andrew~M Dai, Anja Hauth, et~al.
\newblock Gemini: a family of highly capable multimodal models.
\newblock \emph{arXiv preprint arXiv:2312.11805}, 2023.

\bibitem[Wang et~al.(2023)Wang, Pang, and Lin]{wang2023large}
Zhiqiang Wang, Yiran Pang, and Yanbin Lin.
\newblock Large language models are zero-shot text classifiers.
\newblock \emph{arXiv preprint arXiv:2312.01044}, 2023.

\bibitem[Wu and Huang(2022)]{wu2022im2city}
Meiliu Wu and Qunying Huang.
\newblock Im2city: image geo-localization via multi-modal learning.
\newblock In \emph{Proceedings of the 5th ACM SIGSPATIAL International Workshop on AI for Geographic Knowledge Discovery}, pages 50--61, 2022.

\bibitem[Yue et~al.(2023)Yue, Ni, Zhang, Zheng, Liu, Zhang, Stevens, Jiang, Ren, Sun, et~al.]{yue2023mmmu}
Xiang Yue, Yuansheng Ni, Kai Zhang, Tianyu Zheng, Ruoqi Liu, Ge Zhang, Samuel Stevens, Dongfu Jiang, Weiming Ren, Yuxuan Sun, et~al.
\newblock Mmmu: A massive multi-discipline multimodal understanding and reasoning benchmark for expert agi.
\newblock \emph{arXiv preprint arXiv:2311.16502}, 2023.

\bibitem[Zhang et~al.(2022)Zhang, Van~Durme, Li, and Stengel-Eskin]{zhang2022visual}
Chenyu Zhang, Benjamin Van~Durme, Zhuowan Li, and Elias Stengel-Eskin.
\newblock Visual commonsense in pretrained unimodal and multimodal models.
\newblock \emph{arXiv preprint arXiv:2205.01850}, 2022.

\end{thebibliography}
}

% \clearpage

\end{document}